\documentclass[sigconf]{acmart}
\AtBeginDocument{%
  }

\usepackage{tikz}
\usepackage{quantikz}
\usetikzlibrary{positioning}
\usepackage{xcolor}
\usepackage{caption}
\usepackage{subcaption} 
\usepackage{amsmath}
\usepackage{graphicx}
\usepackage{booktabs}
\usepackage{enumitem}
\usepackage{multirow}

\setcopyright{acmlicensed}
\copyrightyear{2018}
\acmYear{2018}
\acmDOI{XXXXXXX.XXXXXXX}
\acmConference[Conference acronym 'XX]{Make sure to enter the correct
  conference title from your rights confirmation email}{June 03--05,
  2018}{Woodstock, NY}
\acmISBN{978-1-4503-XXXX-X/2018/06}

\begin{document}

\title{Investigating Quantum Circuit Designs Using Neuro-Evolution}


\author{Devroop Kar}
\affiliation{%
  \institution{Rochester Institute of technology}
  \city{Rochester NY}
  \country{United States}}
\orcid{0000-0003-1404-1076}
\email{dk7405@rit.edu}

\author{Daniel Krutz}
\affiliation{%
  \institution{Rochester Institute of technology}
  \city{Rochester NY}
  \country{United States}
}
\orcid{0000-0002-0310-3086}
\email{dxkvse@rit.edu}

\author{Travis Desell}
\affiliation{%
  \institution{Rochester Institute of technology}
  \city{Rochester NY}
  \country{United States}
}
\orcid{0000-0002-4082-0439}
\email{tjdvse@rit.edu}





\renewcommand{\shortauthors}{Kar et al.}

\begin{abstract}

Designing effective quantum circuits remains a central challenge in quantum computing, as circuit structure strongly influences expressivity, trainability, and hardware feasibility. Current approaches, whether using manually designed circuit templates, fixed heuristics, or automated rules, face limitations in scalability, flexibility, and adaptability, often producing circuits that are poorly matched to the specific problem or quantum hardware. In this work, we propose the Evolutionary eXploration of Augmenting Quantum Circuits (EXAQC), an evolutionary approach to the automated design and training of parameterized quantum circuits (PQCs) which leverages and extends on strategies from neuroevolution and genetic programming. The proposed method jointly searches over gate types, qubit connectivity, parameterization, and circuit depth while respecting hardware and noise constraints. The method supports both Qiskit and Pennylane libraries, allowing the user to configure every aspect. This work highlights evolutionary search as a critical tool for advancing quantum machine learning and variational quantum algorithms, providing a principled pathway toward scalable, problem-aware, and hardware-efficient quantum circuit design. Preliminary results demonstrate that circuits evolved on classification tasks are able to achieve over 90\% accuracy on most of the benchmark datasets with a limited computational budget, and are able to emulate target circuit quantum states with high fidelity scores.

\end{abstract}




\maketitle

\section{Introduction}\label{sec:introduction}

Quantum machine learning (QML) has emerged as a promising paradigm for leveraging quantum computational principles—such as superposition, entanglement, and interference—to enhance learning and decision-making tasks \cite{chiribella2008quantum,grover1996fast,kitaev2002classical,shor1999polynomial}. Variational quantum circuits (VQCs), which combine parameterized quantum circuits with classical optimization, form the backbone of many contemporary QML approaches \cite{khairy2020learning, lukac2003evolutionary, grimsley2019adaptive} 
. Despite their appeal, the practical success of VQCs depends to a great degree on the architecture, including gate types, connectivity, parameterization, and measurement strategy. Designing such architectures remains a major challenge, particularly as most approaches rely on manually crafted templates or shallow heuristic designs that may not generalize across tasks or datasets.

Circuit structure significantly influences expressivity, trainability, and robustness to noise, often more so than the choice of optimizer or loss function \cite{chen2021expressibility, du2022quantum, hubregtsen2021evaluation}. Poorly chosen architectures may suffer from barren plateaus \cite{pesah2021absence, mcclean2018barren, cerezo2021cost}, weak gradient signals, or insufficient entanglement between input and output qubits. Consequently, there is growing interest in automated quantum circuit discovery methods that jointly learn both circuit topology and parameters rather than fixing the structure beforehand.

Evolutionary computation has demonstrated strong performance in neural architecture search, symbolic regression, and reinforcement learning. Evolutionary algorithms are able to explore large, discrete design spaces without expensive computational overhead. These properties make evolutionary methods particularly well-suited for quantum circuit design, where the search space is inherently combinatorial and involves discrete choices over gates, qubits, and wiring patterns \cite{lukac2003evolutionary,ding2022evolutionary,chen2025evolutionary,zhang2023evolutionary,sunkel2023ga4qco,ruican2008genetic,miranda2021synthesis, kolle2025evaluating}. However, existing quantum architecture search approaches either restrict evolution to shallow templates, or need significant evaluation resources for testing various architectures, limiting their applicability to real-world learning problems.

In this work, we propose an evolutionary framework for task-driven quantum circuit discovery that directly optimizes both circuit structure and parameters for supervised learning on classical datasets as well as emulating specific circuit behavior. Quantum circuits are represented as mutable genomes composed of parameterized and non-parameterized quantum gates, enabling the evolution of circuit depth, gate ordering, qubit connectivity, and entanglement patterns. Circuit parameters are optimized using gradient-based learning, while structural modifications are explored through evolutionary operators, resulting in a hybrid evolutionary--variational training process.

A key feature of the proposed framework is its backend-agnostic design, and is thus able to support both PennyLane and Qiskit execution models. This allows circuits to be trained using statevector simulation, observable-based losses, or probabilistic readout strategies depending on task requirements. Rather than enforcing explicit target quantum states, we employ measurement-driven loss functions, such as cross-entropy over marginal readout probabilities, which naturally align with classical classification objectives. This design enables the direct application of quantum circuits to standard supervised learning benchmarks without requiring artificial quantum labels.

We evaluate the proposed approach on multiple UCI benchmark datasets, including the Iris, Wine, Seeds, and Breast Cancer classification tasks. Classical features are embedded into quantum states via angle-based encodings, while predictions are obtained from designated readout qubits using marginal probability distributions. Through evolution, the framework discovers nontrivial circuit topologies that increasingly entangle input and output registers, leading to improved classification performance. Notably, expressive circuit structures emerge organically from the evolutionary process rather than being imposed through predefined entangling layers.


The contributions of this work are summarized as follows:
\begin{itemize}[leftmargin=*]
    \item We introduce an evolutionary quantum circuit search framework that jointly optimizes circuit structure and parameters for supervised learning tasks.
    \item We present a flexible, backend-agnostic training pipeline compatible with both PennyLane and Qiskit.
    \item We propose measurement-driven loss formulations that enable direct training on classical labeled datasets.
    \item We demonstrate the effectiveness of the approach on multiple real-world benchmark datasets.
    \item We provide practical insights into the role of entanglement, readout design, and circuit structure in quantum model trainability.
\end{itemize}

This work highlights evolutionary architecture search as a practical and scalable methodology for discovering problem-aware quantum circuits.

\section{Related Work}
\label{sec:related_work}

Automated quantum circuit design has recently accelerated, driven by two practical constraints: (I) circuit structure strongly affects trainability, expressivity, and hardware feasibility, and (II) the search space of gate types, connectivity, and depth grows combinatorially. A growing body of work therefore treats quantum circuit synthesis / quantum neural architecture search (QNAS) as an explicit optimization problem, often using evolutionary or neuroevolutionary algorithms that can explore discrete structural choices while optionally refining continuous gate parameters.

A representative line of work focuses on \emph{circuit synthesis and circuit simplification} under fidelity and complexity constraints. S{\"u}nkel et al.~\cite{sunkel2025quantum} propose a hybrid multi-objective evolutionary algorithm that targets (1) maximizing fidelity to a target state and (2) minimizing circuit depth. They evaluate two variants: a purely evolutionary approach (mutation/crossover) and a hybrid approach that periodically applies a classical optimizer to tune rotation parameters; experiments on randomly generated target circuits (4 and 6 qubits) show substantial depth reductions while maintaining high fidelity.
Compared to this direction, our setting emphasizes \emph{dataset-driven training and evaluation} (e.g., supervised classification/regression or task-conditioned objectives), rather than only matching a single target state/circuit under synthetic random targets. We also aim to make the inner-loop training backend-agnostic (PennyLane/Qiskit ML), so that the same evolutionary representation can be evaluated under different execution/training stacks.

Zhang and Zhao~\cite{zhang2022evolutionary} similarly frame QAS as an evolutionary search problem for classification-style objectives, using genetic operators to propose architectures and assessing candidates via learning performance.
Our work differs in how the objective is constructed: instead of comparing only measurement-level metrics, we support \emph{fidelity-based supervision against a target circuit’s input--output state mapping} (when a target circuit is available), while still tracking auxiliary metrics (e.g., angle distance, KL on readout distributions, depth/complexity penalties) to guide search toward trainable and hardware-feasible solutions.

Several other works explicitly adapt \emph{neuroevolution} ideas to variational circuit structure. QNEAT~\cite{giovagnoli2023qneat} introduces a NEAT-inspired mechanism that evolves circuit topology along with parameters, and evaluates on reinforcement-learning-style benchmarks (e.g., FrozenLake and related control tasks), positioning topology growth as a way to balance expressivity and trainability. 
In contrast, our approach uses a unified circuit genome representation that can be evaluated both for task losses and for circuit-to-circuit (or circuit-to-state) similarity, enabling direct fidelity supervision when a known target transformation exists (e.g., small algorithmic primitives), and dataset-driven readout losses otherwise.

Ding and Spector~\cite{ding2023multi} explicitly cast PQC design as \emph{multi-objective} evolutionary search, highlighting that accuracy alone is insufficient and must be balanced against properties such as circuit size and complexity or other surrogate measures of usefulness for near-term devices. This aligns with our design goal: we treat fidelity (or task loss) as necessary but not sufficient, and we incorporate additional metrics during evolution/training to discourage brittle solutions (e.g., depth growth without benefit) and to encourage architectures that actually propagate information from inputs to readouts.

A number of works target supervised classification directly. EQNAS~\cite{li2023eqnas} proposes an evolutionary NAS pipeline for quantum models in image classification settings, using population-based search with evolutionary operators and parameter updates, and reports results on MNIST and domain-specific datasets. 
Li et al.~\cite{li2025aqea} propose an adaptive quantum evolutionary algorithm with elite retention for QAS and evaluate on common vision benchmarks such as MNIST, Fashion-MNIST, and CIFAR-10. 
Ma et al.~\cite{ma2024continuous} present a continuous-evolution approach with multi-objective optimization for efficient QAS, emphasizing trade-offs between predictive performance and circuit complexity. 
Collectively, these works establish evolutionary QNAS as a strong baseline for supervised problems, but they typically couple evaluation to \emph{measurement-space objectives} (accuracy, cross-entropy on readouts) and/or specific templates that are not designed to directly match a target quantum transformation.

Our work is complementary and aims to broaden capability in two ways:
(i) when a target circuit (or a target quantum mapping) exists, we can train/evolve against its \emph{input--output state action} using fidelity-based losses computed over a set of probe inputs (basis states and/or embedded continuous features); and
(ii) when no target circuit exists (standard supervised datasets), we still operate in a consistent pipeline by reducing the model state to readout-wire distributions and computing task losses there, while tracking structural metrics for search guidance.

Structured representations are another recurring theme. Lourens et al.~\cite{lourens2023hierarchical} propose hierarchical circuit representations for NAS and evaluate on tasks spanning classical classification (e.g., GTZAN) and physics-motivated recognition settings, arguing that representation matters for scalable search and meaningful genetic operators. 
Our genome similarly emphasizes modularity (gate specifications, register-aware wiring, and input output-wire designation), but differs in that we explicitly support backend-agnostic execution and a training objective that can switch between readout-space losses and fidelity-to-target mappings.

Not all approaches rely purely on gate-level genetic operators. Ewen et al.~\cite{ewen2025application} explore genetic programming where ZX-calculus-inspired transformations serve as mutation rules, targeting regression function approximation tasks. 
This is related in spirit to our interest in making mutations more ``semantics-aware''. Our framework can incorporate similar ideas by adding structure-preserving mutation operators or rule-guided edits, while retaining the ability to fine-tune continuous parameters with differentiable training.

Hybrid quantum--classical architectures have also been evolved. Liu et al.~\cite{liu2021hybrid} study hybrid quantum--classical convolutional models, demonstrating that search over hybrid building blocks can be beneficial for image tasks. 
Rubio et al.~\cite{rubio2025towards} discuss evolutionary QAS in the context of more fully quantum learning systems, emphasizing efficiency and architectural refinement. 
Our work sits between these directions: we focus on a \emph{general circuit genome} that can represent small algorithmic primitives (for fidelity-to-target training) and dataset-driven classifiers, while supporting strong evaluation signals beyond accuracy alone.

\subsection{Summary and Positioning of our Contribution}
Across these works, two recurring gaps motivate our design:
\begin{enumerate}
  \item \textbf{Limited supervision signal when a target transformation exists.}
  Many QNAS pipelines optimize readout accuracy or a single target state, rather than matching a target circuit's behavior over a set of inputs. We address this by enabling \emph{fidelity-based training against a target circuit’s input--output mapping} (when provided), which creates a denser and more physically grounded objective than classification accuracy alone.
  \item \textbf{Search--training coupling and portability.}
  Existing methods often assume a specific execution/training stack or fixed templates. Our pipeline emphasizes a backend-agnostic circuit representation and training loop, so evolved circuits can be executed and trained in multiple backends without changing the evolutionary machinery.
\end{enumerate}

\section{Methodology}

Evolutionary Exploration of Augmenting Quantum Circuits (EXAQC)\footnote{Source code is open source and publicly available at: \url{https://anonymous.4open.science/r/exaqc-761C/}} takes inspiration from the Evoluationary Exploration of Augmenting Memory Models (EXAMM) neuroevolution algorithm~\cite{ororbia2019examm} and its genetic programming variant Evolutionary Exploration of Augmenting Genetic Programs (EXA-GP)~\cite{murphy2024exa, murphy2024minimizing}. While EXAMM and EXA-GP optimize graph structures which are unconstrained beyond the specified number of input and output nodes, automating the design of quantum circuits is constrained by the number of qubits for the task, where gates are required to be placed on given qubits. Due to this, new mutation operators have been designed for EXAQC.  EXAQC adapts EXAMM and EXA-GP's crossover operator to the quantum circuit paradigm, while additionally adding a new exponential crossover operator as well as a novel \textit{n-ary} crossover operation which extends the binary crossover operator to work on any number of parents.

Given the similarities in its circuit evolution process, EXAQC leverages EXAMM/EXA-GP's Lamarckian weight inheritance~\cite{lyu2021experimental} to reduce the amount of backpropagation required for the evolved circuits. It also uses an distributed, asynchronous master-worker framework to allow multiple quantum circuits to be trained simultaneously.  In its current initial implemenetation, it utilizes a single steady state population, however the codebase has been designed to allow for plugging in varying population strategies so it can be easily extended to utilize island or multiobjective strategies in future work.

\subsection{Genome Representation}
Genomes are represented by a list of input qubit identifiers, a list of output qubit identifiers (which may or may not overlap with the input qubit identifiers) and a list of gates. Each gate has a depth (a float between 0.0 and 1.0), list of input and output qubit identifiers, dictionary of trainable parameters (if it is parameterized) and an innovation number (similar to NEAT~\cite{stanley2002evolving} and EXAMM/EXA-GP) which is is utilized the identify the same gate across multiple genomes for purposes of crossover. Unlike other strategies which utilize a fixed genome size, the list of gates is unbounded allowing for circuits to grow as large as needed.

\begin{figure}
\centering

\begin{subfigure}{0.45\textwidth}
\centering
\begin{quantikz}[row sep=0.25cm, column sep=0.22cm]
\lstick{$q_0$} & \gate{H} & \qw      & \gate{R_z(\theta)} & \qw \\
\lstick{$q_1$} & \qw      & \ctrl{1} & \qw               & \qw \\
\lstick{$q_2$} & \qw      & \targ{}  & \gate{X}          & \qw
\end{quantikz}
\;$\Rightarrow$\;
\begin{quantikz}[row sep=0.25cm, column sep=0.22cm]
\lstick{$q_0$} & \gate{H} & \qw
               & \gate[style={fill=yellow!25,draw=black}]{G_{\text{new}}}
               & \qw & \gate{R_z(\theta)} & \qw \\
\lstick{$q_1$} & \qw & \ctrl{1} & \qw & \qw & \qw & \qw \\
\lstick{$q_2$} & \qw & \targ{}  & \qw & \gate{X} & \qw & \qw
\end{quantikz}
\caption{Add gate}
\label{fig:add_gate_sub}
\end{subfigure}
\vspace{0.75mm}

\begin{subfigure}{0.45\textwidth}
\centering
\begin{quantikz}[row sep=0.25cm, column sep=0.22cm]
\lstick{$q_0$} & \gate{H} & \gate{G_{\text{enabled}}} & \gate{R_z(\theta)} & \qw \\
\lstick{$q_1$} & \qw      & \ctrl{1}           & \qw                & \qw \\
\lstick{$q_2$} & \qw      & \targ{}            & \gate{X}           & \qw
\end{quantikz}
\;$\Leftrightarrow$\;
\begin{quantikz}[row sep=0.25cm, column sep=0.22cm]
\lstick{$q_0$} & \gate{H}
               & \gate[style={fill=yellow!25,draw=black,dashed}]{G_{\text{disabled}}}
               & \gate{R_z(\theta)} & \qw \\
\lstick{$q_1$} & \qw & \ctrl{1} & \qw & \qw \\
\lstick{$q_2$} & \qw & \targ{}  & \gate{X} & \qw
\end{quantikz}
\caption{ Enable/Disable gate}
\label{fig:enable_disable_gate_sub}
\end{subfigure}
\vspace{0.75mm}

\begin{subfigure}{0.45\textwidth}
\centering
\begin{quantikz}[row sep=0.25cm, column sep=0.22cm]
\lstick{$q_0$} & \gate{H}
               & \gate[style={fill=yellow!25,draw=black}]{G_{\text{old position}}}
               & \gate{R_z(\theta)} & \qw \\
\lstick{$q_1$} & \qw & \ctrl{1} & \qw & \qw \\
\lstick{$q_2$} & \qw & \targ{}  & \gate{X} & \qw
\end{quantikz}
\;$\Rightarrow$\;
\begin{quantikz}[row sep=0.25cm, column sep=0.22cm]
\lstick{$q_0$} & \gate{H}
               & \gate{R_z(\theta)}
               & \gate[style={fill=yellow!25,draw=black}]{G_{\text{new position}}} & \qw \\
\lstick{$q_1$} & \qw & \ctrl{1} & \qw & \qw \\
\lstick{$q_2$} & \qw & \targ{}  & \gate{X} & \qw
\end{quantikz}
\caption{Reorder gate}
\label{fig:reorder_gate_sub}
\end{subfigure}
\vspace{0.75mm}

\begin{subfigure}{0.45\textwidth}
\centering
\begin{quantikz}[row sep=0.25cm, column sep=0.22cm]
\lstick{$q_0$} & \gate{H} & \qw                & \qw \\
\lstick{$q_1$} & \qw      & \ctrl{1}          & \qw \\
\lstick{$q_2$} & \qw      & \targ{}           & \gate{G} \\
\lstick{$q_3$} & \qw      & \qw               & \qw
\end{quantikz}
\;$\Rightarrow$\;
\begin{quantikz}[row sep=0.25cm, column sep=0.22cm]
\lstick{$q_0$} & \gate{H} & \qw                & \qw \\
\lstick{$q_1$} & \qw      & \qw                & \qw \\
\lstick{$q_2$} & \qw
               & \targ{}
               & \gate[style={fill=yellow!25,draw=black}]{G} \\
\lstick{$q_3$} & \qw
               & \ctrl{-1}
               & \qw
\end{quantikz}
\caption{Qubit swap}
\label{fig:qubit_swap_sub}
\end{subfigure}

\caption{Structural mutation operators used in hybrid evolutionary quantum circuit optimization. Highlighted gates indicate newly added or modified structures.}
\label{fig:mutation_ops_grid}
\end{figure}
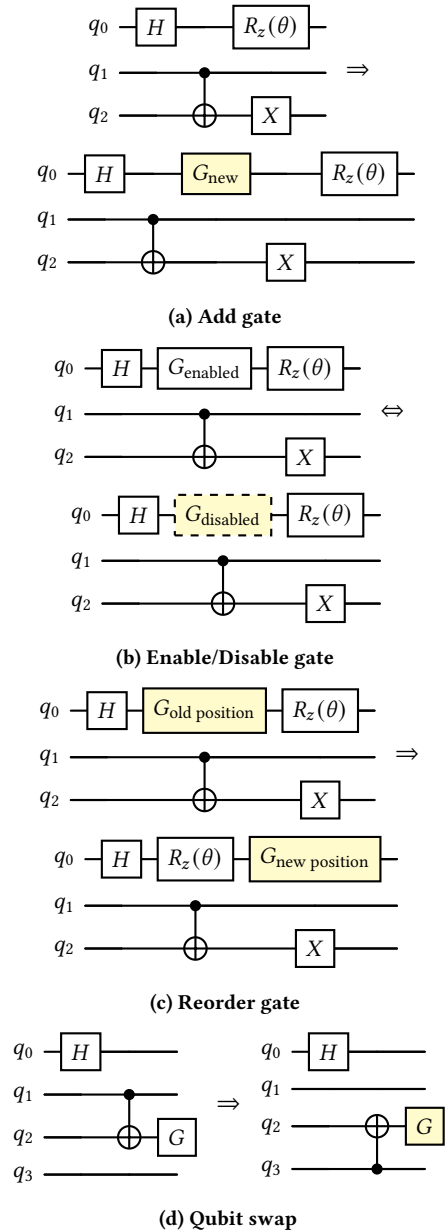

\subsection{Mutation Operators}

Figure~\ref{fig:mutation_ops_grid} presents visual representations of each of the mutation operations used by EXAQC. These mutations are used to gradually increase and refine the size of the quantum circuits.  Note that it is possible for the \textit{disable gate}, \textit{reorder gate} and \textit{swap qubits} mutations to result in circuits where the inputs do not effect the outputs.  After a genome is created by mutation (or crossover), a forward traversal is performed over its gates in sorted order to make sure that at least one input flows to at least one output. If this is not the case, the genome is discarded as invalid and EXAQC will attempt to generate another one via mutation or crossover.

\subsubsection{Add Gate}
EXAQC provides a wide range of potential gates (selected uniformly at random with some constraints described below, which depending on the circuit connectivity at the depth the gate is being added), which span nearly the entire range of available gates in Qiskit or Pennylane. For Pennylane, 23 gates are available for selection, and 43 gate are available for Qiskit. In general, all gates are available for use from Qiskit except for some multi-controlled gates, which can be designed by using other gates or applying certain operators on specific gates. 
Fewer Pennylane gates are available as it does not implement as wide a range as Qiskit. Appendixes~\ref{app:qiskit_gates} and~\ref{app:pennylane_gates} in the supplementary material provide the lists of available gates in EXAQC for Qiskit and Pennylane, respectively.

The add gate mutation (Figure~\ref{fig:add_gate_sub}) first determines a random depth, $d$, to place the gate, determined uniformly at random, $d \sim U(0,1)$. As the input and output qubits of a circuit can be disjoint, to prevent gates which will not be involved in useful computation to be added, the gates in the network are traversed in order from 0 to $d$, to determine which qubits are effected by the inputs - these make the list of potential input qubits for the new gate.  Similarly, the gates are traversed in reverse order from $1.0$ to $d$, to determine which qubits are connected to the outputs by other gates -- these make the list of potential output qubits for the new gate. If the list of input and output qubits are disjoint, only gates which have control (input) and target (output) qubits are allowed. The input qubits for the new gate are selected from the possible input qubits, and the output qubits for the new gate are selected from the possible output qubits.  If the gate is parameterized, its parameters are initialized randomly $\sim U(-\pi,\pi)$. The parent circuit genome will be copied as the new child, and the gate will be added to the child genome at depth $d$ with a newly assigned innovation number.

\subsubsection{Enable/Disable Gate}
The enable and disable gate mutations (Figure~\ref{fig:enable_disable_gate_sub}) first copies the parent as a child genome, then randomly selects from the child either an enabled gate to be disabled (for \textit{disable gate}), or a disabled gate to be enabled (for \textit{enable gate}). If \textit{enable gate} is selected and there are no disabled gates, or \textit{disable gate} is selected and there are no enabled gates, this mutation will return False so another mutation or crossover attempt can be made to generate a child.

\subsubsection{Reorder Gate}
Reorder gate (Figure~\ref{fig:reorder_gate_sub}) first copies the parent as a child genome, then randomly selects a gate within the child genome. It disables that gate, creates a copy of it (with a new innovation number), and assigns it a new depth $d \sim U(0,1)$, a new innovation number, and adds it to the child genome.

\subsubsection{Swap Qubits}
Swap qubits (Figure~\ref{fig:qubit_swap_sub}) first copies the parent as a child genome, then randomly selects a gate within the child genome. It disables that gate, creates a copy of it (with a new innovation number), and then selects a qubit parameter at random with the gate and assigns it to a different qubit. If the qubit was an input qubit, the new qubit is randomly selected from all possible input qubits at that depth, if it was an output qubit, the new qubit is randomly selected from all possible output qubits at that depth. Otherwise (if it was both an input and output) it is randomly selected from all possible input and output qubits. Additionally, to prevent the new gate from having the same depth as the gate it was copied from, it is assigned a new depth $d \sim U(d_{prev},d_{next})$, where $d_{prev}$ is the depth of the previous gate in the (sorted) gate list, or 0 if there was not a gate before it, and $d_{max}$ is the depth of the next gate in the gate list, or 1.0 if there was not a gate after it.  This new gate is assigned a new innovation number and added to the child genome.

\subsection{Crossover}

EXAQC adapts the binary crossover methodology from EXAMM and EXA-GP, while also expanding it to allow for $n$-ary crossover with any number of parents ($n=4$ for this work).  Additionally, as the gates are stored in an ordered list, it is also possible to utilize an exponential crossover strategy.

\subsubsection{Binary Crossover}
In binary crossover (Figure~\ref{fig:binary_crossover_sub}) a child genome is created and all the gates which are present in both parents are copied into the child genome. If a gate exists in the more fit parent, but not the less fit parent, it is added at a specified $best\_keep\_rate$ (0.75 for this work). If a gate exists in the less fit parent, but not in the more fit parent, it is added at a specified $other\_keep\_rate$ (0.25 for this work). For the parameterized gates that are in both parents, each new weight, $p_{new}$ is computed by recombining the two parent weights, $p_{best}$ and $p_{other}$ utilizing a randomized line search, where $l_1=-1.0$ and $l_2=0.5$ are used as hyperparameters:
\begin{equation} \label{eq1}
\begin{aligned}
r=(rand(0,1)*l_{1})-l_{2} \\
p_{new}=p_{other}+r*(p_{best}-p_{other})
\end{aligned}
\end{equation}
\noindent
Gates which only came from one parent utilize the same weights. This follows EXAMM/EXA-GP's Lamarckian weight inheritance strategy~\cite{lyu2021experimental}.

\subsubsection{N-Ary Crossover}

N-ary crossover (Figure~\ref{fig:n_ary_crossover_sub}) expands binary crossover to allow any number of parents. Parents are divided into the best fit parent, and all other parents. If a gate exists in the best fit parent and any other parent, it is added to the child genome at the $best\_keep\_rate$. If a gate exists in any other gate, but not the best fit parent, it is added to the child genome at the $other\_keep\_rate$. For parameterized gates in multiple parents, instead of a randomized line search, a randomized simplex approach is used~\cite{kini2023co,thakur2023efficient}, where the randomized line search is performed between weight in the best parent, $p_{best}$ and the average of the weights in the other parents $p_{avg}$, with $r$ generated the same as in binary crossover:
\begin{equation}\label{eq3}
p_{new}=p_{avg}+r*(p_{best}-p_{avg})
\end{equation}

\subsubsection{Exponential Crossover}
Exponential crossover (Figure~\ref{fig:exponential_crossover_sub}) was also implemented as a way to recombine parents in a way that could potentially capture different useful gates from each parent. This creates a new child genome, and computes a random crossover depth $d_{crossover} \sim U(0,1)$. All gates from the first parent with depth $d < d_{crossover}$ are copied into the child genome, and all gates from the second parent with depth $d \geq d_{crossover}$ are copied from the second parent into the child genome.

\providecommand{\bothgate}[1]{\fcolorbox{black}{yellow!25}{\strut #1}}
\providecommand{\bestgate}[1]{\fcolorbox{black}{blue!12}{\strut #1}}
\providecommand{\othergate}[1]{\fcolorbox{black}{orange!15}{\strut #1}}

\providecommand{\sharedgate}[1]{\fcolorbox{black}{yellow!25}{\strut #1}}
\providecommand{\primaryonly}[1]{\fcolorbox{black}{blue!12}{\strut #1}}
\providecommand{\otheronly}[1]{\fcolorbox{black}{orange!15}{\strut #1}}

\begin{figure}[!h]
\centering
\scriptsize

\textbf{Parent 1 (best)}\par\vspace{-0.2em}
\begin{quantikz}[row sep=0.18cm, column sep=0.18cm]
\lstick{$q_0$} & \gate{\bestgate{$i{=}1\,H$}}
               & \gate{\bothgate{$\mathrm{CNOT}$}}
               & \gate{\bothgate{$R_z(\theta)$}}
               & \gate{\bestgate{$R_x(\alpha)$}} & \qw \\
\lstick{$q_1$} & \qw & \ctrl{1} & \qw & \qw & \qw \\
\lstick{$q_2$} & \qw & \targ{}  & \qw & \qw & \qw
\end{quantikz}

\vspace{0.9em}

\textbf{Parent 2 (other)}\par\vspace{-0.2em}
\begin{quantikz}[row sep=0.18cm, column sep=0.18cm]
\lstick{$q_0$} & \qw
               & \gate{\bothgate{$\mathrm{CNOT}$}}
               & \gate{\bothgate{$R_z(\theta)$}}
               & \gate{\othergate{$X$}}
               & \gate{\othergate{$R_y(\beta)$}} & \qw \\
\lstick{$q_1$} & \qw & \ctrl{1} & \qw & \qw & \qw & \qw \\
\lstick{$q_2$} & \qw & \targ{}  & \qw & \qw & \qw & \qw
\end{quantikz}

\vspace{1.0em}
{\Large$\Downarrow\;\texttt{Binary Crossover}\;\Downarrow$}
\vspace{1.2em}

\textbf{Child (recombined)}\par\vspace{-0.2em}
\begin{quantikz}[row sep=0.18cm, column sep=0.18cm]
\lstick{$q_0$} & \gate{\bestgate{$H$}}
               & \gate{\bothgate{$\mathrm{CNOT}$}}
               & \gate{\bothgate{$R_z(\theta_{\text{child}})$}}
               & \gate{\othergate{$X$}}
               & \gate{\othergate{$R_y(\beta)$}}
               & \qw \\
\lstick{$q_1$} & \qw & \ctrl{1} & \qw & \qw & \qw & \qw \\
\lstick{$q_2$} & \qw & \targ{}  & \qw & \qw & \qw & \qw
\end{quantikz}

\vspace{0.8em}

\begin{minipage}{0.98\linewidth}
\scriptsize
\textbf{Selection rules:}
Shared innovations (\bothgate{yellow}) always transfer.
Best-only innovations (\bestgate{blue}) transfer with probability $p_b$.
Other-only innovations (\othergate{orange}) transfer with probability $p_o$.\par
\textbf{Parameter recombination (shared, parameterized gates):}
$r \leftarrow U(0,1)(c_2-c_1)+c_1$,\;
$\theta_{\text{child}}=\theta_{p2}+r(\theta_{p1}-\theta_{p2})$.
\end{minipage}

\caption{Binary crossover operator for innovation-numbered circuit genomes.}
\label{fig:binary_crossover_sub}
\end{figure}
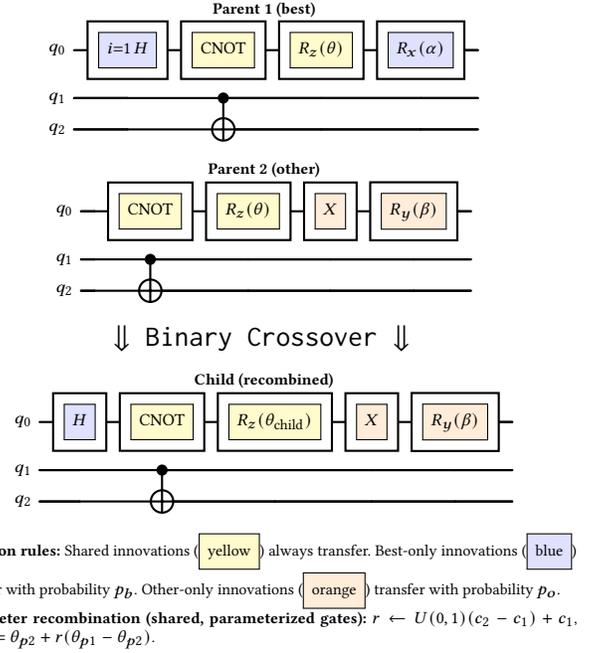


\begin{figure}[!h]
\centering
\scriptsize

\textbf{Parent 1}\par\vspace{-0.2em}
\begin{quantikz}[row sep=0.16cm, column sep=0.18cm]
\lstick{$q_0$} & \gate{\bestgate{$G_1$}} & \gate{\bestgate{$G_2$}} & \gate{\bestgate{$G_3$}} & \qw \\
\lstick{$q_1$} & \qw                    & \ctrl{1}                & \qw                     & \qw \\
\lstick{$q_2$} & \qw                    & \targ{}                 & \qw                     & \qw
\end{quantikz}
\par\vspace{-0.1em}
{\color{black}\textbf{keep gates with depth $< d_c$}}

\vspace{0.9em}

\textbf{Parent 2}\par\vspace{-0.2em}
\begin{quantikz}[row sep=0.16cm, column sep=0.18cm]
\lstick{$q_0$} & \gate{\othergate{$H_1$}} & \gate{\othergate{$H_2$}} & \gate{\othergate{$H_3$}} & \qw \\
\lstick{$q_1$} & \qw                     & \ctrl{1}                 & \qw                      & \qw \\
\lstick{$q_2$} & \qw                     & \targ{}                  & \qw                      & \qw
\end{quantikz}
\par\vspace{-0.1em}
{\color{black}\textbf{keep gates with depth $\ge d_c$}}

\vspace{1.0em}
{\Large$\Downarrow\;\texttt{Exponential Crossover}(d_c)\;\Downarrow$}
\vspace{1.2em}

\textbf{Child (recombined)}\par\vspace{-0.2em}
\begin{quantikz}[row sep=0.16cm, column sep=0.18cm]
\lstick{$q_0$} & \gate{\bestgate{$G_1$}} & \gate{\bestgate{$G_2$}} & \gate{\othergate{$H_3$}} & \qw \\
\lstick{$q_1$} & \qw                    & \ctrl{1}                & \qw                      & \qw \\
\lstick{$q_2$} & \qw                    & \targ{}                 & \qw                      & \qw
\end{quantikz}

\vspace{0.8em}

\begin{minipage}{0.98\linewidth}
\scriptsize
\textbf{Rule:} Sample $d_c \sim U(0,1)$. Copy all gates from Parent~1 with depth $< d_c$,
and all gates from Parent~2 with depth $\ge d_c$, into the child.
\end{minipage}

\caption{Exponential crossover operator.}
\label{fig:exponential_crossover_sub}
\end{figure}
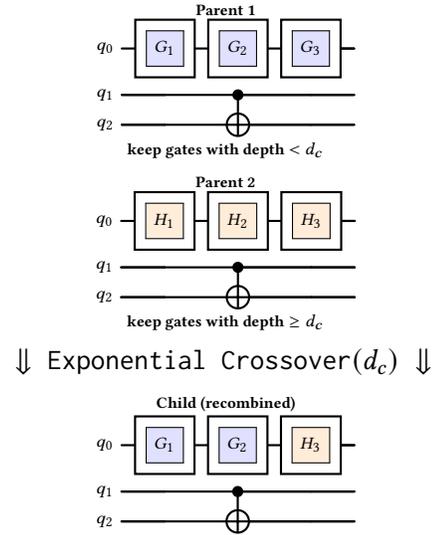


\begin{figure}[!h]
\centering
\scriptsize

\textbf{Primary parent }($P_0$)\par\vspace{-0.2em}
\begin{quantikz}[row sep=0.16cm, column sep=0.16cm]
\lstick{$q_0$} & \gate{\primaryonly{$H$}}
               & \gate{\sharedgate{$\mathrm{CNOT}$}}
               & \gate{\sharedgate{$R_z(\theta)$}}
               & \gate{\primaryonly{$R_x(\alpha)$}} & \qw \\
\lstick{$q_1$} & \qw & \ctrl{1} & \qw & \qw & \qw \\
\lstick{$q_2$} & \qw & \targ{}  & \qw & \qw & \qw
\end{quantikz}

\vspace{0.9em}

\textbf{Other parent }($P_1$)\par\vspace{-0.2em}
\begin{quantikz}[row sep=0.16cm, column sep=0.16cm]
\lstick{$q_0$} & \qw
               & \gate{\sharedgate{$\mathrm{CNOT}$}}
               & \gate{\sharedgate{$R_z(\theta)$}}
               & \gate{\otheronly{$X$}} & \qw \\
\lstick{$q_1$} & \qw & \ctrl{1} & \qw & \qw & \qw \\
\lstick{$q_2$} & \qw & \targ{}  & \qw & \qw & \qw
\end{quantikz}

\vspace{0.9em}

\textbf{Other parent }($P_2$)\par\vspace{-0.2em}
\begin{quantikz}[row sep=0.16cm, column sep=0.16cm]
\lstick{$q_0$} & \qw
               & \gate{\sharedgate{$\mathrm{CNOT}$}}
               & \gate{\sharedgate{$R_z(\theta)$}}
               & \gate{\otheronly{$R_y(\beta)$}} & \qw \\
\lstick{$q_1$} & \qw & \ctrl{1} & \qw & \qw & \qw \\
\lstick{$q_2$} & \qw & \targ{}  & \qw & \qw & \qw
\end{quantikz}

\vspace{1.0em}
{\Large$\Downarrow\;\texttt{N-ary Crossover}(r)\;\Downarrow$}
\vspace{1.2em}

\textbf{Child (recombined)}\par\vspace{-0.2em}
\begin{quantikz}[row sep=0.16cm, column sep=0.16cm]
\lstick{$q_0$} & \gate{\primaryonly{$H$}}
               & \gate{\sharedgate{$\mathrm{CNOT}$}}
               & \gate{\sharedgate{$R_z(\theta_{\text{child}})$}}
               & \gate{\otheronly{$X$}}
               & \gate{\otheronly{$R_y(\beta_{\text{avg}})$}} & \qw \\
\lstick{$q_1$} & \qw & \ctrl{1} & \qw & \qw & \qw & \qw \\
\lstick{$q_2$} & \qw & \targ{}  & \qw & \qw & \qw & \qw
\end{quantikz}

\vspace{0.8em}

\begin{minipage}{0.98\linewidth}
\scriptsize
\textbf{Selection rules:}
If innovation $i$ appears in the primary parent and in at least one other parent (\sharedgate{yellow}), it always transfers to the child.
If $i$ appears only in the primary parent (\primaryonly{blue}), it transfers with probability $p_{\mathbf{p}}$.
If $i$ appears only in the other parents (\otheronly{orange}), it transfers with probability $p_o$.\par

\textbf{Parameter recombination (shared, parameterized gates):}
$r \leftarrow U(0,1)(c_2-c_1)+c_1$, \;
$\overline{\theta}_{\text{others}}=\frac{1}{|\mathcal{O}_i|}\sum_{k\in\mathcal{O}_i}\theta_k$, \;
$\theta_{\text{child}}=\theta_{\text{o}} + r(\overline{\theta}_{\mathbf{p}}-\theta_{\text{o}})$.\par

\textbf{Other-only parameter averaging (when multiple others):}
For a selected other-only innovation with parameters, set $\beta_{\text{avg}}=\frac{1}{m}\sum_{j=1}^{m}\beta_j$.
\end{minipage}

\caption{N-ary crossover operator.}
\label{fig:n_ary_crossover_sub}
\end{figure}

\subsection{Loss Functions for Evolutionary Quantum Circuit Optimization}
\label{sec:losses}

Evaluating candidate quantum circuits during evolutionary search requires objective functions that are informative, differentiable under simulation, and aligned with the intended task. In this work, we employ multiple loss functions to support two complementary optimization settings: (i) imitation of a target (teacher) quantum circuit and (ii) supervised classification using quantum readout probabilities. Each loss induces a different fitness landscape, influencing genome search and selection behavior.

\subsubsection{Fidelity-Based State Loss}
\label{subsec:fidelity}

For circuit to circuit comparison, candidate circuits are evaluated by comparing their output quantum states against a fixed teacher circuit. Given a predicted state $\ket{\psi}$ and target state $\ket{\phi}$, fidelity score is defined as:
\begin{equation}
F(\phi, \psi) = |\langle \phi \mid \psi \rangle|^2 
\end{equation}

The fidelity score, can also be expressed in terms of density matrices $\rho$ and $\sigma$,

\begin{equation}
F(\rho, \sigma) = \Biggl( tr\sqrt{\sqrt{\rho}\sigma\sqrt{\rho}}\Biggr)^2
\end{equation}

A fidelity score of 1.0 represents complete overlap between 2 quantum states. In practical scenarios a score of 0.9 to 0.99 is preferred. The corresponding loss is conceived as:
\begin{equation}
\mathcal{L}_{\text{fid}} = 1 - F(\phi, \psi).
\end{equation}

\textbf{Role in evolution.} Fidelity provides a smooth, bounded fitness signal that directly measures quantum state similarity. It enables structure discovery without relying on classical labels and is well-suited for evolving circuits that reproduce entanglement, interference, and multi-qubit correlations.

\textbf{Scope.} This loss requires access to full statevectors and is therefore used in simulation-based experiments.

\subsubsection{Angular State Distance}
\label{subsec:angle}

As an alternative geometric measure, we use the angular distance between quantum states,
\begin{equation}
\mathcal{L}_{\text{angle}} = \arccos\left(|\langle \phi \mid \psi \rangle|\right).
\end{equation}

\textbf{Role in evolution.} Unlike fidelity, angular distance maintains gradient sensitivity near high overlap, which can reduce premature convergence when candidate circuits approach the teacher behavior.

\subsubsection{Distribution-Based Losses}
\label{subsec:distribution}

When optimization is based on measurement statistics rather than full states, we compare probability distributions derived from quantum amplitudes.

\paragraph{Kullback--Leibler Divergence.}
Given $p_i = |\psi_i|^2$ and $q_i = |\phi_i|^2$, the KL divergence is
\begin{equation}
\mathcal{L}_{\text{KL}} = \sum_i q_i \log \frac{q_i}{p_i}.
\end{equation}

\textbf{Role in evolution.} KL divergence emphasizes mismatches in high-probability measurement outcomes, encouraging correct observable statistics even when underlying quantum states differ in phase.

\paragraph{Observable Mean Squared Error.}
For hardware-aligned settings, we define an observable loss:
\begin{equation}
\mathcal{L}_{\text{MSE}} =
\frac{1}{K} \sum_{k=1}^{K}
\left(
\langle O_k \rangle_{\psi} - \langle O_k \rangle_{\phi}
\right)^2 ,
\end{equation}
where $O_k$ are Pauli observables on designated output qubits.

\textbf{Role in evolution.} Observable-based losses restrict feedback to quantities measurable on quantum hardware, enabling evolutionary optimization under realistic constraints.

\subsubsection{Cross-Entropy on Quantum Readout Probabilities}
\label{subsec:ce}

For supervised classification, circuits output measurement probabilities over a designated readout register. Let $p \in \mathbb{R}^K$ denote the marginal distribution mapped to $K$ classes and $y$ the one-hot encoded label. The cross-entropy loss is:
\begin{equation}
\mathcal{L}_{\text{CE}} = - \sum_{k=1}^{K} y_k \log(p_k).
\end{equation}

\textbf{Role in evolution.} Cross-entropy directly aligns quantum outputs with classical learning objectives and provides strong selection pressure for discriminative behavior. Probabilities are renormalized over class-relevant basis states to accommodate limited readout qubits.

\subsubsection{Loss Function Summary}

The framework supports multiple loss functions, enabling controlled experiments on how fitness definitions influence evolutionary dynamics:
\begin{itemize}
    \item \textbf{Teacher imitation:} fidelity, angular distance
    \item \textbf{Measurement-driven optimization:} KL divergence, observable MSE
    \item \textbf{Supervised learning:} cross-entropy on readout probabilities
\end{itemize}

\section{Experimental Setup} \label{sec:experimental_setup}

We evaluate the proposed evolutionary framework using both classical supervised datasets and synthetic teacher quantum circuits. This combination allows us to study generalization on labeled data and structural learning in purely quantum settings.

\subsection{Benchmark Problems: Datasets and Qauntum Circuits}
\label{sec:benchmarks}

\subsubsection{Classical Supervised Datasets}
\label{subsec:datasets}

We employ several standard UCI datasets frequently used in quantum machine learning research. These datasets span increasing input dimensionality and task complexity, allowing systematic analysis of circuit expressivity and parameter efficiency under evolutionary search. The datasets selected were:
\begin{itemize}
\item \textbf{Iris:} Four real-valued features and three classes. This dataset provides a low-dimensional baseline for rapid evolutionary search.
\item \textbf{Wine:} Thirteen features and three classes. This benchmark evaluates scalability of quantum encodings and circuit depth.
\item \textbf{Seeds:} Seven geometric features and three classes. This dataset presents moderately complex class boundaries.
\item \textbf{Breast Cancer:} Thirty features with binary labels. This benchmark tests high-dimensional input handling and robustness under class imbalance.
\end{itemize}

\subsubsection{Teacher Quantum Circuits}
\label{subsec:teacher}

To isolate architectural learning from classical data effects, we randomly selected fixed teacher quantum circuits that define target input–output behavior. Candidate circuits are evolved to minimize state-level losses relative to the teacher. These benchmarks were utilized as teacher circuits provide a fully quantum benchmark where success depends on reproducing entanglement structure and interference patterns rather than fitting noisy labels. The teacher circuits used as benchmarks were:
\begin{itemize}
    \item Identity and single-gate transformations (baseline validation) - These included single qubit transformations using gates like Identity, Hadamard, Pauli-X etc.
    \item Bell-state generators (entanglement learning) - These created simple entanglement using Hadamard and CNOT.
    \item Input-controlled output circuits (cross-register information flow) - These included further entanglements using CNOT, Toffoli etc. across input and output wires.
    \item Multi-layer fixed circuits with rotations and entangling gates - These circuits contained combination of parameterized and standard gates together with entanglements.
\end{itemize}

\subsubsection{Why Both Benchmarks Are Necessary}

Supervised datasets evaluate the framework as a quantum classifier, while teacher circuits assess its ability to discover quantum circuit structure independent of classical supervision. Together, these benchmarks demonstrate that the proposed evolutionary approach supports both task-driven and physics-driven quantum circuit optimization.
\section{Experimental Results}\label{sec:experimental_results}

All EXAQC experiments utilized a single steady state population with a maximum size of 50 genomes. Pennylane was used as the target quantum framework. Individuals were only generated by mutation given a base empty genome (no gates) until the population was full. When mutation was performed, it utilized \textit{add gate} at 70\%, \textit{reorder gate} at 10\%, \textit{qubit swap} at 10\%, \textit{enable gate} at 5\%, and \textit{disable gate} at 5\%. After the population was initialized, \textit{binary crossover} was performed at 10\%, \textit{n-ary crossover} (with $n=4$) at 10\%, \textit{exponential crossover} at 10\%, and mutation (as above) at 70\%. When mutation was selected (both pre- and post-initialization), two mutations were performed on the genome to speed the evolution process. Each experiment was run until a total of 500 genomes were evaluated, using 12 processes (1 master and 11 workers). For training parameterized gates, the networks were trained for 200 epochs using the Adam optimizer with a learning rate of 0.001 and weight decay of 0.0001. 


\subsection{Classification Benchmarks}

\begin{table}
\begin{tabular}{lccc}
\hline
 \textbf{Dataset} & \textbf{Test Acc. \%} & \textbf{\# Gates} & \textbf{Genome \#} \\
\hline
\multirow{5}{*}{Iris}   & 86.7\%    & 21    & 735 \\
                        & 70.0\%    & 7     & 592 \\
                        & 83.3\%    & 12    & 754 \\
                        & 70.0\%    & 12    & 683 \\
                        & 90.0\%    & 21    & 800 \\

\hline
\multirow{5}{*}{Seeds}  & 92.9\%    & 12    & 790 \\
                        & 92.9\%    & 8     & 888 \\
                        & 95.2\%    & 13    & 787 \\
                        & 90.5\%    & 12    & 898\\
                        & 92.9\%    & 8     & 733 \\

\hline
\multirow{5}{*}{Wine}   & 75.0\%    & 11    & 625 \\
                        & 75.0\%    & 15    & 804 \\
                        & 77.8\%    & 12    & 876 \\
                        & 88.9\%    & 16    & 851 \\
                        & 86.1\%    & 15    & 734 \\
\hline
                        & 91.2\%    & 13    & 849 \\
Breast                  & 89.2\%    & 10    & 765 \\
Cancer                  & 90.8\%    & 11    & 735 \\
                        & 88.6\%    & 11    & 860 \\
                        & 91.0\%    & 12    & 831 \\
\hline
\end{tabular}
\caption{Experimental results for each classification benchmark.}
\label{tab:class_results}
\end{table}
The evolved quantum circuits were trained on the classification datasets using cross-entropy loss, utilizing the readout probabilities for output quantum states against each of the class labels. We used 6$\sim$8 qubits as input and $\lceil{\log_2n}\rceil$ as the number of outputs where $n$ is the number of classes. Results across the benchmark datasets are provided in Table \ref{tab:class_results}. 

Even with a very modest budget of 500 evaluated genomes, we find that EXAQC can quickly find well performing quantum circuits for classification tasks. The genome number column\footnote{Note that as time mutation was performed it was done twice, which incremented the genome number twice; which is why the genome numbers are higher than 500. Additionally, if a genome was generated and discarded for not having a path from inputs to outputs, that would also increment the genome counter.} shows which genome found the best performing circuit. Across the benchmark datasets, we find that the largest circuits were the best performing - which suggests evolving circuits for even longer could provide even better results.  Additionally, we find that many the lower performing experiments had their best genome found earlier in the search, which may suggest they have stalled out. Appendix~\ref{app:best_classification_circuits} in the supplementary material also provides diagrams of the best found circuits for each benchmark. We note that for the best wine benchmark circuit, two qubits were not connected to the measured output qubits, and in the best breast cancer circuit, three qubits are not connected to the measured output qubits which also suggests better performance could be obtained with a longer search.

\subsection{Teacher Quantum Benchmarks}
We also performed experiments which utilized the quantum ``teacher'' circuits which generated a sample input and output state space. We evolved our quantum circuits to learn the behavior and configuration of the teacher circuits. The \textit{Fidelity} between the target and our circuit outputs helped to drive the evolution process. The final fidelity and angular distance measures between the teacher circuit output states for these experiments and the output from the evolved circuits have been reported in Table \ref{tab:fid_results}.

\begin{table}[!h]
  \centering
  \caption{Fidelity and Angular Distance measure comparison against teacher circuits.}
  \label{tab:fid_results}
  \setlength{\tabcolsep}{7pt} 
  \renewcommand{\arraystretch}{1.15} 

  \begin{tabular}{lccc}
    \toprule
    \textbf{Circuit}  & \textbf{Fidelity} & \textbf{Angular Distance} \\
    \midrule
    Baseline & 100.00\% & 0.0 \\
    Bell-State Generator  & 98.32\% & 0.014 \\
    Input Controlled & 94.11\% & 0.058 \\
    Multi-Layered & 91.73\% & 0.101 \\
    \bottomrule
  \end{tabular}
\end{table}

On the baseline circuits which only compare single gate transformations, our evolved circuits were able to emulate the teacher circuit very quickly, within 3 to 5 genomes, and all reached a fidelity of 100\% and angular distance of 0.0. For Bell State generators, our evolved circuits captured the target states with high fidelity and with a low amount genome exploration. For input controlled gates where the chosen input and output qubits were highly entangled, the evolution process still managed to capture the target states to a high degree but required more genome explorations. For more multi-layered complex circuits the process required over 800 genomes.

\subsection{Discussion}

From Table \ref{tab:class_results} it can be noted that across all these datasets the evolved networks are able to achieve good classification results, with relatively few generated genomes -- if the strategy was synchronous and population based, these results would have been obtained in less than 10 generations. We also note that performance could be further improved by modifying the importance of each mutation operator and gate across different generation of genomes is important and can lead to further better results. Initially, the \textit{add gate} would be the most important to make sure the evolved circuit connects the inputs and outputs appropriately, as evolution begins, there is need for adding gates rather than disabling or even reordering them; but with successive exploration we may focus more on refining the better genomes to improve quantum state representations.

S{\"u}nkel et al. \cite{sunkel2025quantum} uses a similar evolutionary strategy but with mutation and crossover operators having different implementations. Their search process is restricted to fixed set of 5 gates and their evaluation process focuses only on synthetic circuits. Our evolutionary setup addresses these shortcoming by implementing nearly all quantum gate specifications as supplied by Pennylane and/or Qiskit, and does not restrict itself to any specific list of gates. Our methodology does provide the option to specify a search should focus on specific gates, otherwise it searches through all of them. For future work, dynamically adapting which gates are selected based on how much benefit they provide to the circuit could be beneficial.

\begin{figure}
    \centering
    \includegraphics[scale=0.275]{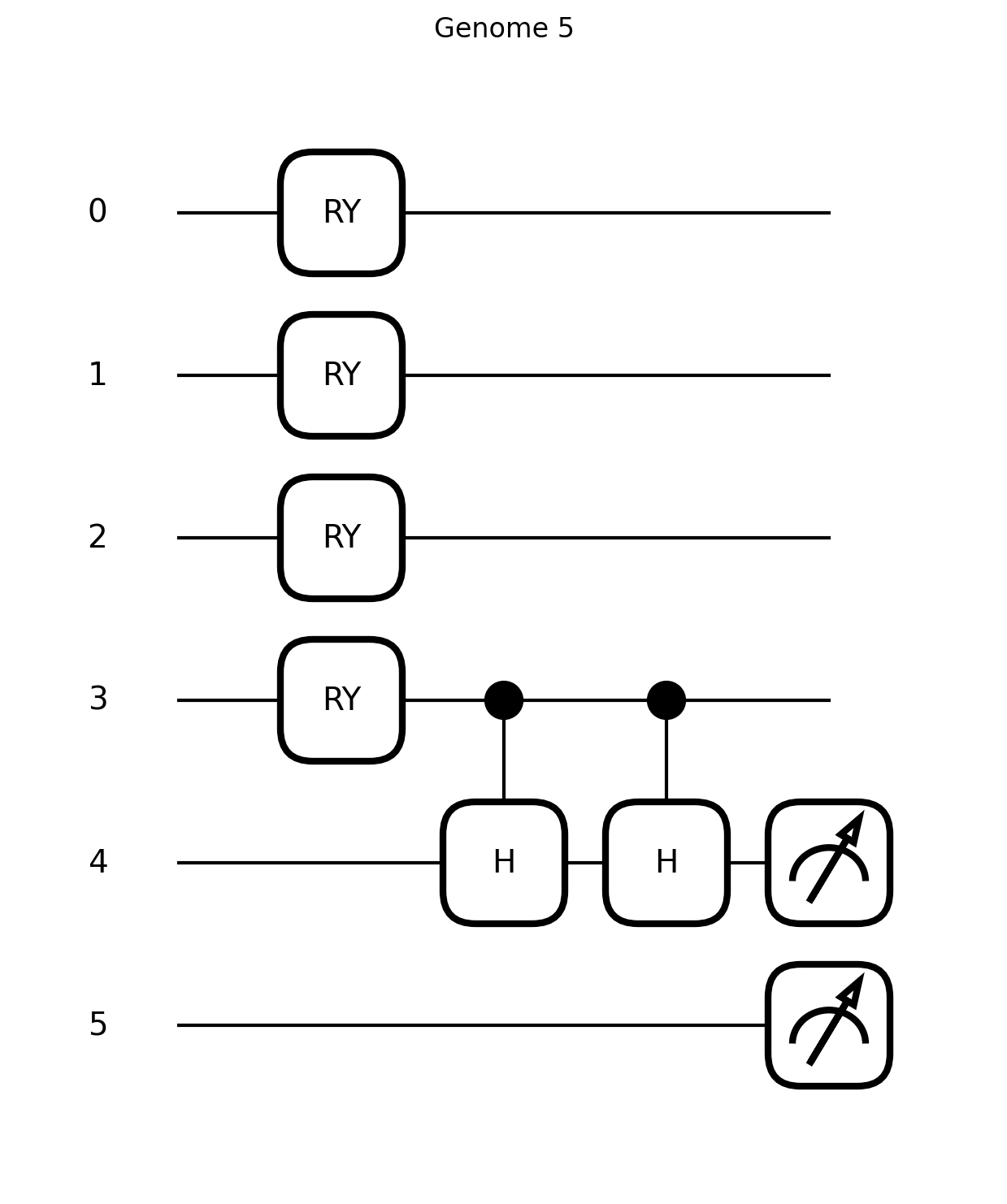}
    \hfill
    \includegraphics[scale=0.275]{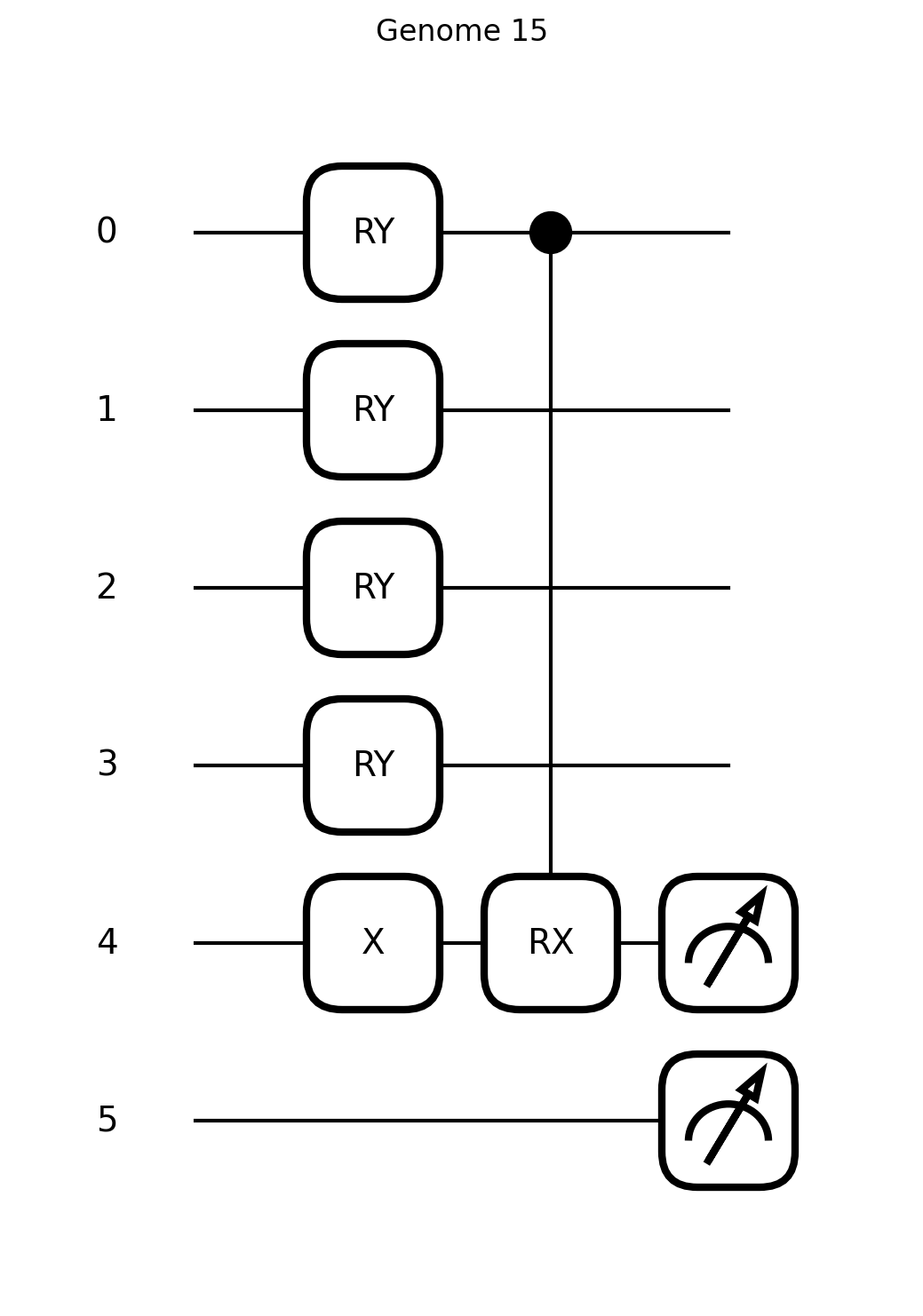}
    \caption{An example baseline identity gate as learned by two different genomes.}
    \label{fig:evolve_identity}
\end{figure}

EXAQC also is generic enough to be repurposed to evolve circuits across different domains. Table \ref{tab:fid_results} highlights how our process is able identify behavioral traits of the teacher circuits and evolve circuits that can capture their state representation with high fidelity. Figures \ref{fig:evolve_identity} represents 2 genomes which reached a 100\% fidelity score against the teacher circuits but were structurally different from the teacher circuits. The first genome attempted to learn an identity gate on qubits 4 and 5. As seen while qubit 5 was untouched, the Hadamard gate was applied twice on qubit 4 - thus effectively acting like an identify gate. In the second case a Pauli-X gate was applied to only qubit 4. The evolved circuit was able to correctly identify the transformation, but it further applied a $R_\mathbf{X}(\theta=0)$ gate, to make it act as an indentity gate.

\section{Conclusion}\label{sec:conclusion}

In this paper, we propose the Evolutionary eXploration of Augmenting Quantum Circuits (EXAQC) algorithm to design and train quantum circuits. Our methodology is generic, and can evolve circuits for both the Qiskit and Pennylane quantum frameworks. It is parallelized for improved performance and leverages advanced crossover and Lamarckian weight inheritance methodolgies adapted from neuroevolution and genetic programming strategies. Our results demonstrate that the evolved circuits can be learn to solve classical classification problems with improving accuracy and even generate quantum circuits to emulate pre-conceived circuits and their input output quantum with high fidelity. This framework is also not limited to any fixed set of gates and supports nearly all gates as supported by prevalent quantum computing libraries. 

Results show the promise of the EXAQC methodology and lay the groundwork for significant future work. EXAQC allows for any quantum objective function to be plugged into the search process, which can allow circuits can be trained to potentially model any optimization problem. Expanding the library to include examples for reinforcement learning, time series forecasting and more complicated classification tasks (such as computer vision) will highlight its ability as a generic quantum circuit optimization framework. Further, EXAQC currently only utilizes a single steady state population and utilizes a single objective for optimization.  Future work will involve investigating utilizing multiple islands and different speciation strategies to enhance optimization, as well as adapting these population strategies to incorporate multi-objective optimization to better utilize the variety of loss metrics EXAQC utilizes.


\balance
\bibliographystyle{ACM-Reference-Format}
\bibliography{refs}

\appendix
\clearpage
\onecolumn
\section{Appendix}

\subsection{Available Qiskit Gates}
\label{app:qiskit_gates}
\begin{table}[h!]
\begin{tabular}{llll}
\hline
\textbf{Gate} & \textbf{Method} & \textbf{Qubits} & \textbf{Parameters} \\
\hline
Toffoli & ccx & control\_qubit1, control\_qubit2, target\_qubit &  \\
\hline
Symmetric & ccz & control\_qubit1, control\_qubit2, target\_qubit &  \\
\hline
Controlled Hadamard & ch & control\_qubit, target\_qubit &  \\
\hline
Controlled Phase & cp & control\_qubit, target\_qubit & theta \\
\hline
Controlled RX & crx & control\_qubit, target\_qubit & theta \\
\hline
Controlled RY & cry & control\_qubit, target\_qubit & theta \\
\hline
Controlled RZ & crz & control\_qubit, target\_qubit & theta \\
\hline
Controlled S & cs & control\_qubit, target\_qubit &  \\
\hline
Controlled S\textsuperscript{\textdagger} & csdg & control\_qubit, target\_qubit &  \\
\hline
Controlled SWAP (Fredkin) & cswap & control\_qubit, target\_qubit1, target\_qubit2 &  \\
\hline
Controlled sqrt X & csx & control\_qubit, target\_qubit &  \\
\hline
Controlled U & cu & control\_qubit, target\_qubit & theta, phi, lam, gamma \\
\hline
Controlled X & cx & control\_qubit, target\_qubit &  \\
\hline
Controlled Y & cy & control\_qubit, target\_qubit &  \\
\hline
Controlled Z & cz & control\_qubit, target\_qubit &  \\
\hline
Double CNOT & dcx & qubit1, qubit2 &  \\
\hline
Echoed Cross-Resonance & ecr & qubit1, qubit2 &  \\
\hline
Hadamard & h & qubit &  \\
\hline
Identity & id & qubit &  \\
\hline
iSWAP & iswap & qubit1, qubit2 &  \\
\hline
Phase & p & qubit & theta \\
\hline
R & r & qubit & theta, phi \\
\hline
Simplified 3-Controlled Toffoli & rcccx & control\_qubit1, control\_qubit2, control\_qubit3, target\_qubit &  \\
\hline
Simplified Toffoli (Margolus) & rccx & control\_qubit1, control\_qubit2, target\_qubit &  \\
\hline
RV & rv & qubit & vx, vy, vz \\
\hline
RX & rx & qubit & theta \\
\hline
RXX & rxx & qubit1, qubit2 & theta \\
\hline
RY & ry & qubit & theta \\
\hline
RYY & ryy & qubit1, qubit2 & theta \\
\hline
RZ & rz & qubit & phi \\
\hline
RZX & rzx & qubit1, qubit2 & theta \\
\hline
RZZ & rzz & qubit1, qubit2 & theta \\
\hline
S & s & qubit &  \\
\hline
S-adjoint & sdg & qubit &  \\
\hline
SWAP & swap & qubit1, qubit2 &  \\
\hline
sqrt X & sx & qubit &  \\
\hline
inverse sqrt X & sxdg & qubit &  \\
\hline
T & t & qubit &  \\
\hline
T-adjoint & tdg & qubit &  \\
\hline
U & u & qubit & theta, phi, lam \\
\hline
X & x & qubit &  \\
\hline
Y & y & qubit &  \\
\hline
z & z & qubit &  \\
\hline
\hline
\end{tabular}
\caption{Available Qiskit gates, their qubits and parameters (if parameterized).}
\label{tab:qiskit_gates}
\end{table}

\clearpage
\subsection{Available Pennylane Gates}
\label{app:pennylane_gates}

\begin{table}[h!]
\begin{tabular}{llll}
\hline
\textbf{Gate} & \textbf{Method} & \textbf{Qubits} & \textbf{Parameters} \\
\hline
Toffoli & ccx & control\_qubit1, control\_qubit2, target\_qubit &  \\
\hline
CCZ & ccz & control\_qubit1, control\_qubit2, target\_qubit &  \\
\hline
Controlled Hadamard & ch & control\_qubit, target\_qubit &  \\
\hline
Controlled Phase & cp & control\_qubit, target\_qubit & phi \\
\hline
Controlled RX & crx & control\_qubit, target\_qubit & phi \\
\hline
Controlled RY & cry & control\_qubit, target\_qubit & phi \\
\hline
Controlled RZ & crz & control\_qubit, target\_qubit & phi \\
\hline
Controlled SWAP (Fredkin) & cswap & control\_qubit, target\_qubit1, target\_qubit2 &  \\
\hline
Controlled X & cx & control\_qubit, target\_qubit &  \\
\hline
Controlled Y & cy & control\_qubit, target\_qubit &  \\
\hline
Controlled Z & cz & control\_qubit, target\_qubit &  \\
\hline
Hadamard & h & qubit &  \\
\hline
Identity & id & qubit &  \\
\hline
iSWAP & iswap & qubit1, qubit2 &  \\
\hline
Phase & p & qubit & phi \\
\hline
RX & rx & qubit & phi \\
\hline
RY & ry & qubit & phi \\
\hline
RZ & rz & qubit & phi \\
\hline
RZZ & rzz & qubit1, qubit2 & theta \\
\hline
S & s & qubit &  \\
\hline
SWAP & swap & qubit1, qubit2 &  \\
\hline
T & t & qubit &  \\
\hline
U & u & qubit & theta, phi, delta \\
\hline
X & x & qubit &  \\
\hline
Y & y & qubit &  \\
\hline
Z & z & qubit &  \\
\hline
\hline
\end{tabular}
\caption{Available Pennylane gates, their qubits and parameters (if parameterized).}
\label{tab:pennylane_gates}
\end{table}

\clearpage
\subsection{Best Found Classification Circuits}
\label{app:best_classification_circuits}

\begin{figure}[!h]
    \centering
    \includegraphics[width=1.0\linewidth]{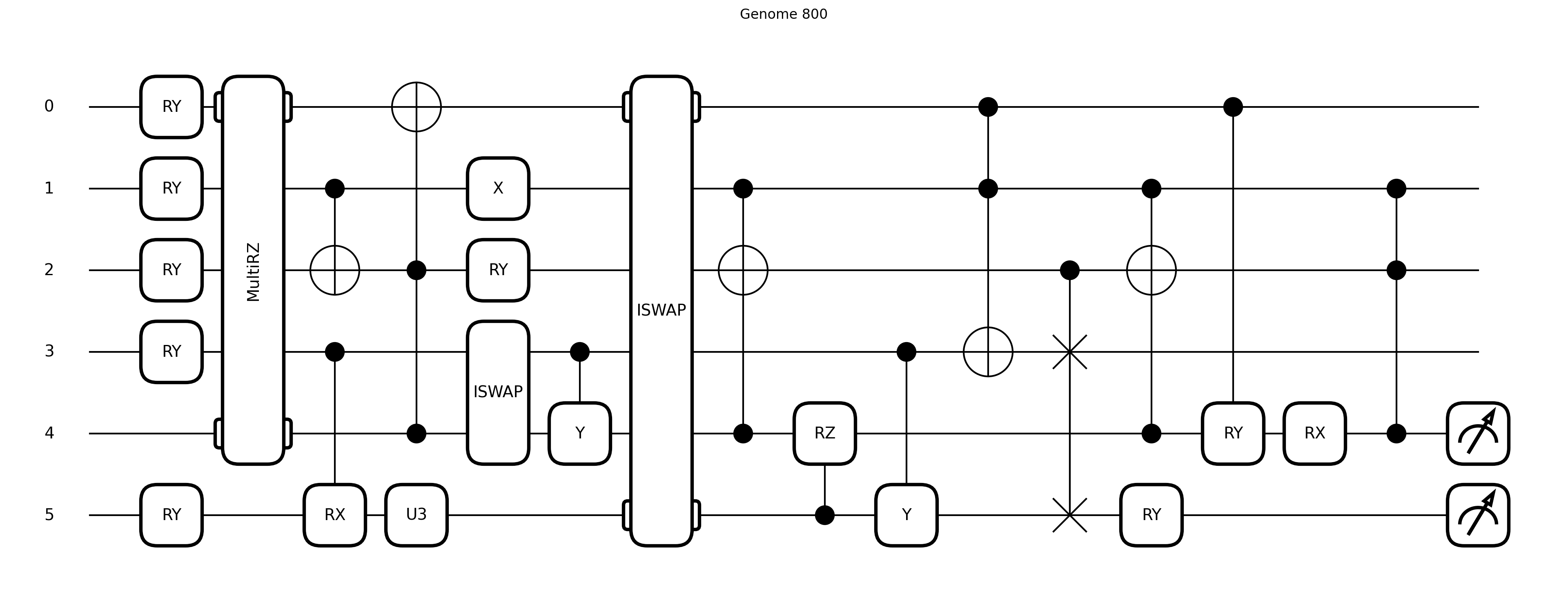}
    \caption{Best found circuit for the iris benchmark.}
    \label{fig:best_iris_circuit}
\end{figure}

\begin{figure}[!h]
    \centering
    \includegraphics[width=0.9\linewidth]{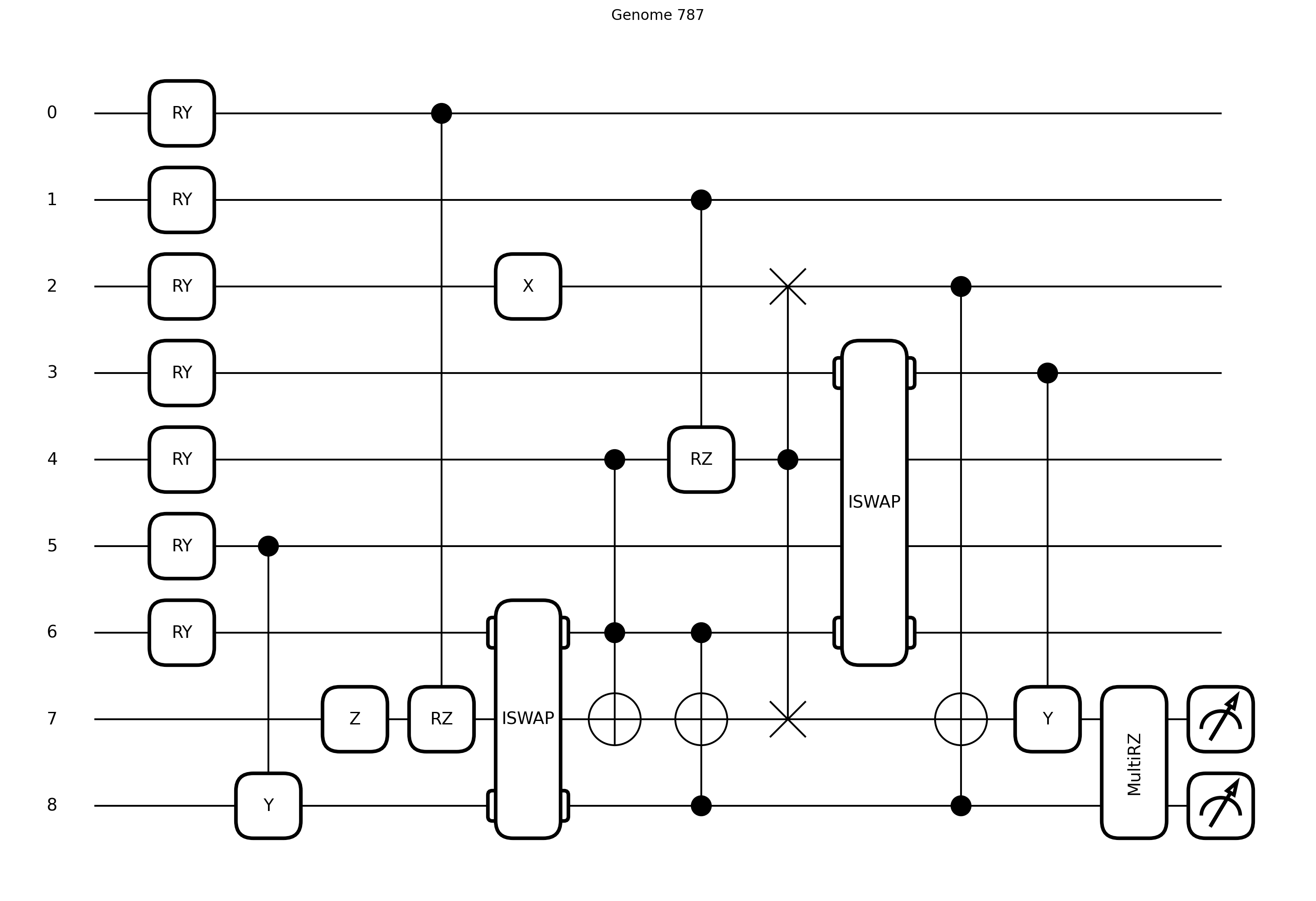}
    \caption{Best found circuit for the seeds benchmark.}
    \label{fig:best_seeds_circuit}
\end{figure}

\begin{figure}[!h]
    \centering
    \includegraphics[width=0.8\linewidth]{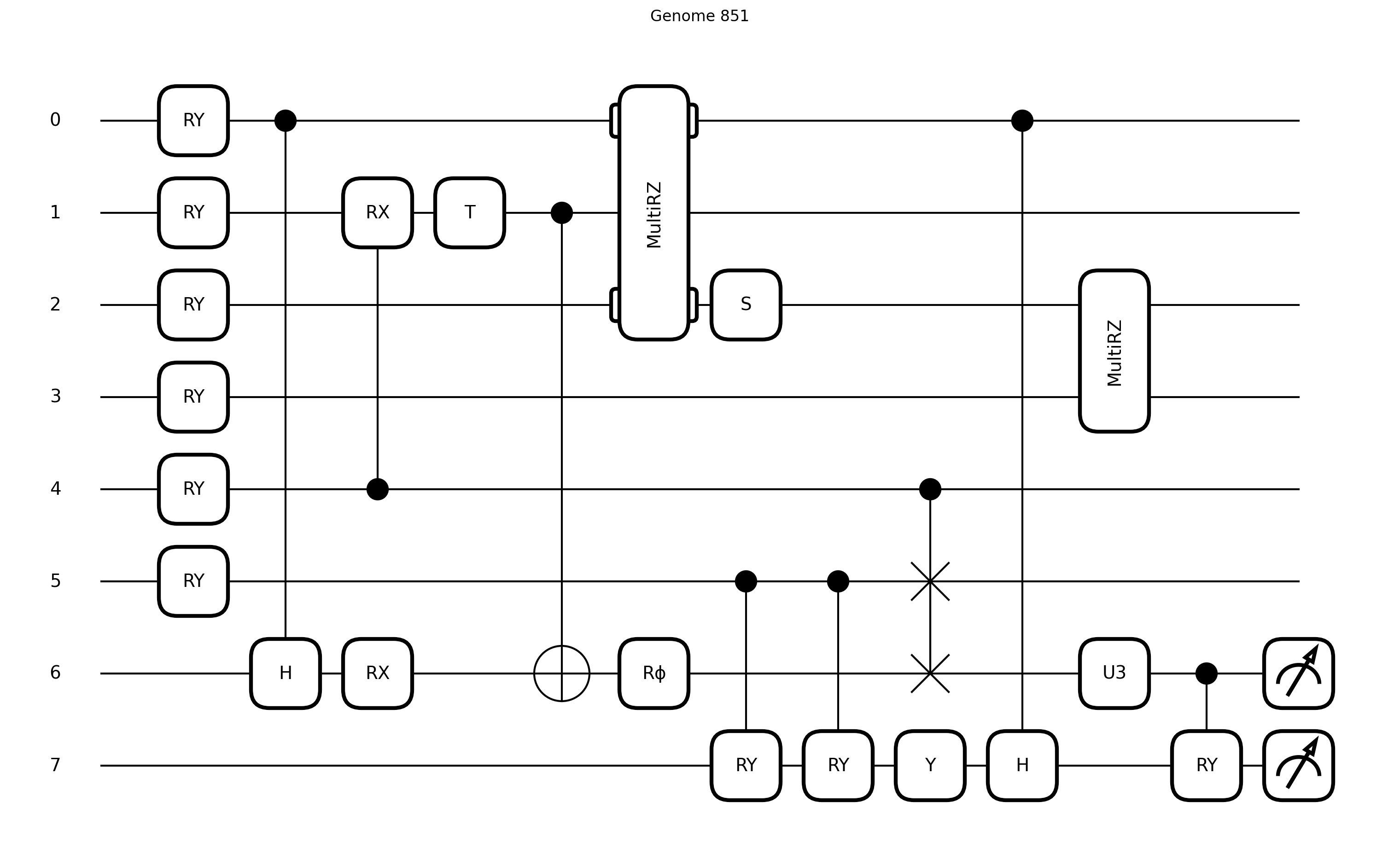}
    \caption{Best found circuit for the wine benchmark.}
    \label{fig:best_wine_circuit}
\end{figure}

\begin{figure}[!b]
    \centering
    \includegraphics[width=0.8\linewidth]{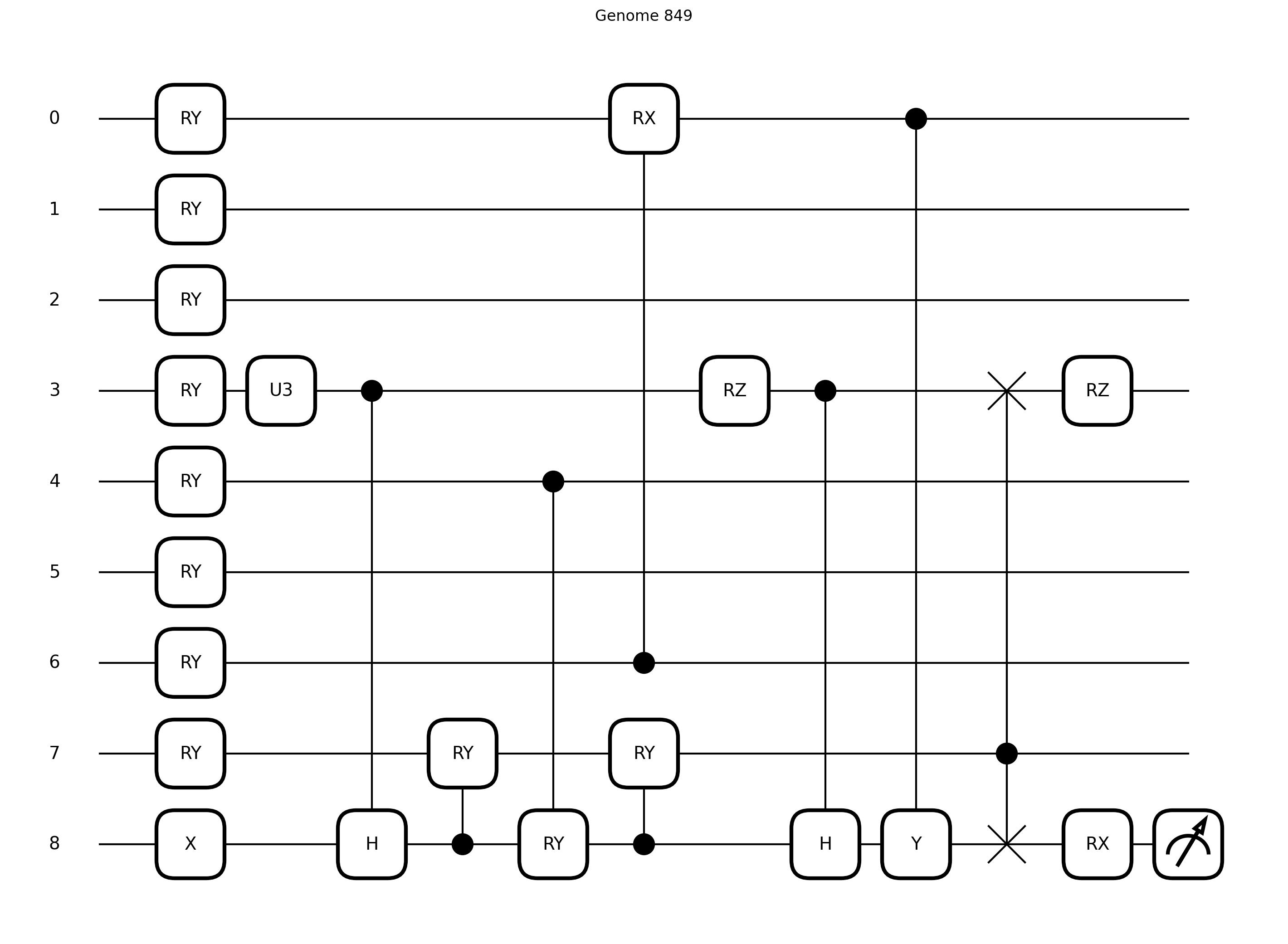}
    \caption{Best found circuit for the breast cancer benchmark.}
    \label{fig:best_breast_cancer_circuit}
\end{figure}

\end{document}